\documentclass[letterpaper, 10 pt, conference]{ieeeconf}  

\IEEEoverridecommandlockouts                              

\usepackage{tabularx}
\usepackage{adjustbox}
\usepackage{graphicx}
\usepackage{fancyvrb}

\usepackage{booktabs}
\usepackage{multirow}
\usepackage{float}
\usepackage{svg}
\let\labelindent\relax
\usepackage{enumitem}
\usepackage[font=small]{caption}

\usepackage{amsmath,amssymb,amsfonts}
\usepackage{graphicx}
\usepackage{textcomp}
\usepackage{xcolor}
\usepackage{booktabs} 
\usepackage{float}
\usepackage{algorithm,algorithmicx,algpseudocode}
\usepackage{multirow}
\usepackage{enumitem}
\usepackage{mathtools}

\usepackage{enumitem}
\usepackage{tikz}
\usepackage{fancyhdr}

\newcommand*\circled[1]{\tikz[baseline=(char.base)]{
            \node[shape=circle,fill,inner sep=1pt] (char) {\textcolor{white}{#1}};}}

\title{\LARGE \bf
\vspace{20pt}
DECADE: Towards Designing Efficient-yet-Accurate Distance Estimation Modules for Collision Avoidance in \\ Mobile Advanced Driver Assistance Systems}

\author{Muhammad Zaeem Shahzad, Muhammad Abdullah Hanif, and Muhammad Shafique
\thanks{Muhammad Zaeem Shahzad, Muhammad Abdullah Hanif, and Muhammad Shafique are with the eBRAIN Lab, New York University Abu Dhabi (NYUAD), UAE.
        {\tt\small \{ms12297, mh6117, muhammad.shafique\}@nyu.edu}}%
}

\begin{document}

\maketitle
\thispagestyle{fancy}
\fancyhf{}
\chead{© 2024 IEEE. This is the author’s version of the work. The definitive Version of Record will be Published in \\ the 2024 IEEE/RSJ International Conference on Intelligent Robots and Systems (IROS).}

\begin{abstract}

The proliferation of smartphones and other mobile devices provides a unique opportunity to make Advanced Driver Assistance Systems (ADAS) accessible to everyone in the form of an application empowered by low-cost Machine/Deep Learning (ML/DL) models to enhance road safety. For the critical feature of Collision Avoidance in Mobile ADAS, lightweight Deep Neural Networks (DNN) for object detection exist, but conventional pixel-wise depth/distance estimation DNNs are vastly more computationally expensive making them unsuitable for a real-time application on resource-constrained devices. In this paper, we present a distance estimation model, DECADE, that processes each detector output instead of constructing pixel-wise depth/disparity maps. In it, we propose a pose estimation DNN to estimate allocentric orientation of detections to supplement the distance estimation DNN in its prediction of distance using bounding box features. We demonstrate that these modules can be attached to any detector to extend object detection with fast distance estimation. Evaluation of the proposed modules with attachment to and fine-tuning on the outputs of the YOLO object detector on the KITTI 3D Object Detection dataset achieves state-of-the-art performance with 1.38 meters in Mean Absolute Error and 7.3\% in Mean Relative Error in the distance range of 0-150 meters. Our extensive evaluation scheme not only evaluates class-wise performance, but also evaluates range-wise accuracy especially in the critical range of 0-70m.

\end{abstract}

\section{INTRODUCTION}

Road traffic accidents are currently among the primary contributors to deaths and injuries on a global scale, resulting in over 1.2 million deaths annually, and are predicted to become the 7th leading cause of death across all age groups globally by 2030 \textcolor{green}{~\cite{who}}. To comply with safety standards, the automotive industry has dedicated its efforts and finances to the development of Advanced Driver Assistance Systems (ADAS) that maximize driver safety by avoiding road traffic accidents. These ADAS are extremely robust systems for real-time, high performance driver facilitation as they employ complex state-of-the-art embedded systems based on high precision sensors like LiDAR.

Since 90\% of the global road traffic accident deaths occur in low and middle-income countries \textcolor{green}{~\cite{who}}, there is a need to further reduce cost and increase accessibility of this life-saving technology to the lower socio-economic classes. \textit{With increasing smartphone adoption rates and the advent of machine learning (ML) algorithms \textcolor{green}{~\cite{survey, survey2}}, a low-cost alternative to the high-end embedded systems for ADAS can be delivered in a Mobile ADAS application by employing advanced deep neural network (DNN) modules that rely on visual input from the smartphone’s camera}. In this way, a Mobile ADAS can facilitate safe driving in non-AI featured low-end automobiles \textcolor{green}{~\cite{resource}}.

\begin{figure}[t]
\centering
\includegraphics[width=1\linewidth]{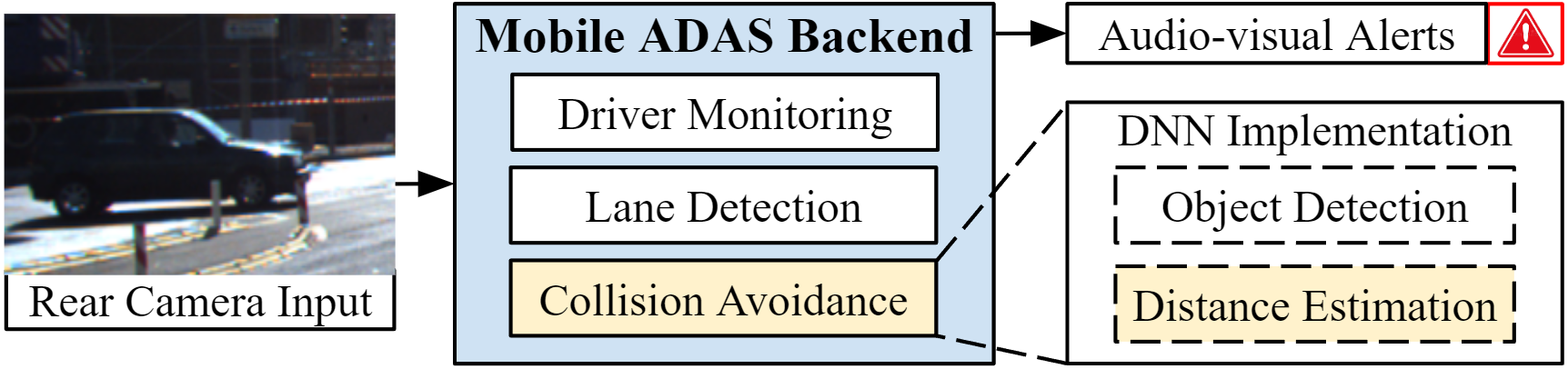}
\caption{Overview of Our Mobile Advanced Driver-Assistance System (ADAS)~\vspace{-5pt}}
\label{fig:3}
\end{figure}

\begin{figure}[!t]
\centering
\includegraphics[width=1\linewidth]{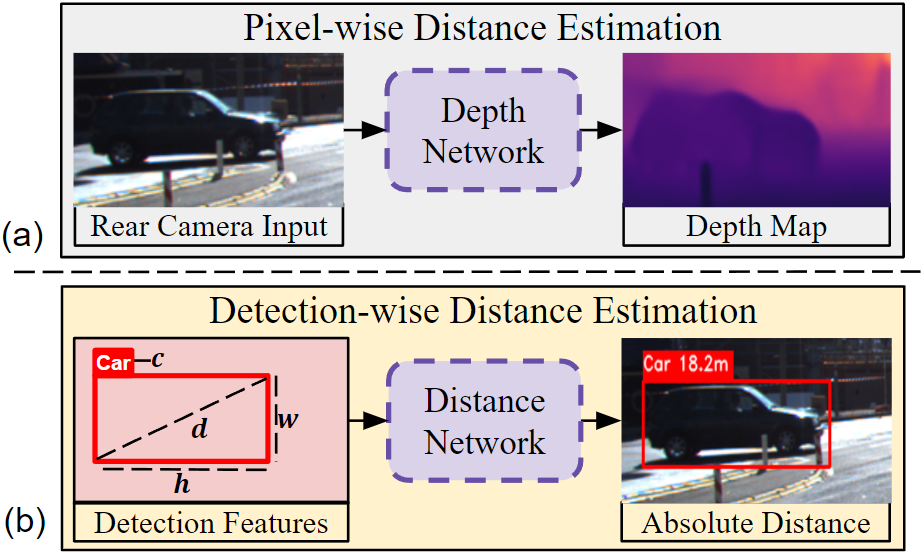}
\caption{Pixel-wise vs Detection-wise Distance Estimation. (a) presents pixel-wise distance estimation where the input image is processed without detector involvement. (b) presents detection-wise distance estimation where only the features extracted from the detector's outputs are processed.~\vspace{-15pt}}
\label{fig:3}
\end{figure}

In this paper, we concentrate on a DNN-based implementation of the critical feature of collision avoidance in ADAS, highlighted in Figure~1. Collision avoidance aims to alert the driver of dangerous scenarios that may result in a collision by monitoring the road scene ahead of the car. The implementation of collision avoidance involves the use of specialized DNNs to solve the problems of object detection and distance estimation. An object detector regresses the bounding box around detected objects and identifies object classes in an input image. Subsequently, the distance estimation network predicts the absolute distance of each detection from the camera. 

It is important to note that there are two distinct classes of distance estimation networks: pixel-wise and detection-wise regressors (see Figure 2). Firstly, in pixel-wise regression (Figure 2a), the input image is processed to generate a pixel-wise depth/disparity map, without detector's involvement. Subsequently, using the detector's output, the distance of each detection can be extracted by post-processing the pixels enclosed by its bounding box \textcolor{green}{~\cite{detdisconventional}}. Secondly, detection-wise regressors (Figure~2b) only process features extracted from the detector's outputs to assign an absolute distance prediction to each detection. 

\begin{figure}[!t]
\centering
\includegraphics[width=1\linewidth]{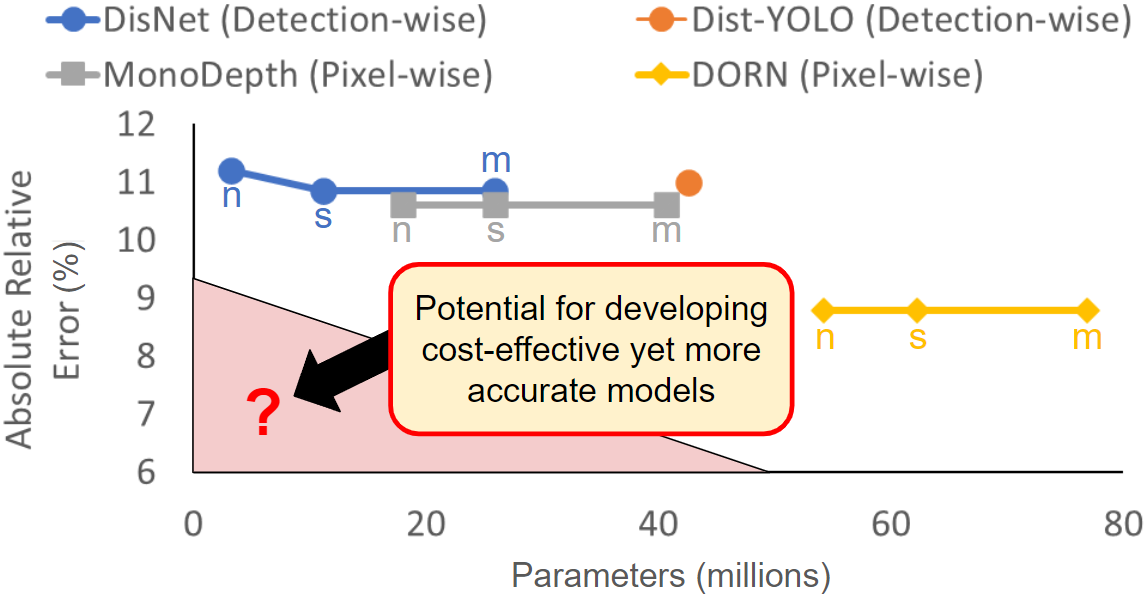}
\caption{Distance estimation accuracy vs. parametric complexity of state-of-the-art collision avoidance DL-implementation modules for mobile ADAS. The labeled distance estimation models are combined with the YOLOv8 n/s/m object detection variants for a fair comparison across pixel-wise and detection-wise domains, except for the Dist-YOLO in which distance estimation is embedded into the YOLOv3 architecture.\vspace{-15pt}}
\label{figmotivational}
\end{figure}

The DNNs used in the mobile ADAS should offer high quality real-time performance while adhering to the hardware constraints of smartphones. Popular object detectors like the You-Only-Look-Once (YOLO) \textcolor{green}{~\cite{yolov3,yolov7}} models have inspired the development of low-cost detectors that provide state-of-the-art performance specifically for low-end hardware like smartphones\textcolor{green}{~\cite{ultralytics}}. However, pixel-wise distance estimation models are vastly more computationally expensive as they regress depth/disparity for each pixel in the image \textcolor{green}{~\cite{monodepth, dorn, densedepth}}. This makes them impractical for a real-time application. However, since a system for collision avoidance applications already employs a detector, DNNs for distance estimation can derive useful information from the detector outputs to predict detection-wise absolute distance instead of processing the image separately. This idea is based off the observation that closer objects appear larger and vice versa. In this way, reducing the scope of the problem to regressing over all input image pixels to only regressing over the number of detections made by the detector, detection-wise distance regression reduces DNN complexity greatly. 

Dist-YOLO \textcolor{green}{~\cite{distyolo}} does this by integrating distance estimation directly into the YOLOv3 architecture. The authors extend the prediction heads of YOLOv3 to make predictions of absolute distance based on the same feature maps learned to make high quality detections. While this approach yields reasonable accuracy across distance metrics, the effect of detector size on the accuracy is unclear. On the other hand, the DisNet \textcolor{green}{~\cite{disnet}} utilizes an additional Multi-Layer Perceptron (MLP) to predict absolute distance for each detection based on features extracted from the outputs of YOLOv3. This approach provides an attachable DNN to extend detectors with distance estimation. Furthermore, since such a DNN is not deeply embedded into a specific Detector like the Dist-YOLO, it can be attached to more lightweight detectors like the YOLOv8 nano (n), small (s), and medium (m) sized variants \textcolor{green}{~\cite{ultralytics}} to reduce overall system cost even further.

\subsection{Motivational Case Study}
Figure~3 visualizes the need for designing efficient-yet-accurate detection-wise distance estimation models. For the most impactful state-of-the-art works in depth/distance estimation including the Dist-YOLO, DisNet, MonoDepth2, and DORN, we compare the distance estimation accuracy, in terms of absolute relative error, with the overall system complexity of the distance network combined with the YOLOv8 object detection variants, in terms of total number of trainable parameters \textcolor{green}{\cite{distyolo, disnet, monodepth, dorn}}. 
\textit{This study illustrates that there is potential to further improve the accuracy of the most cost-effective distance estimation models to offer state-of-the-art performance at low costs.}

\subsection{Our Novel Contributions}
To achieve a better accuracy-efficiency tradeoff for distance estimation, in this paper, we present our DECADE model which extends object detectors with a detection-wise distance estimation network, supplemented by a pose estimation network to infer objects' rotation in the road scene to enhance distance estimation accuracy. Our proposed model can be fine-tuned on detector outputs to push beyond the state of the art in accuracy with 7.3\% absolute relative error while retaining the cost-effectiveness of the most efficient models. Details on the experimental setup can be found in Evaluation and Discussion.
In summary, we make the following key contributions.
\begin{enumerate}[leftmargin=*]
    \item DECADE: A new state-of-the-art attachable distance estimation model consisting of two neural networks. For each detection, a pose estimation network derives orientation information to supplement the rest of the extracted dimensional and spatial features in processing by a distance estimation network. Ultimately, DECADE achieves a new frontier in detection-wise distance estimation accuracy.
    \item Class-wise evaluation: We examine the performance of both DECADE and DisNet across the different object classes in the KITTI dataset to evaluate robustness of distance estimation against variation in object sizes.
    \item Range-wise evaluation: As performance in the distance ranges of 0-70m is critical to driver safety, we perform a detailed range-wise evaluation of both our DECADEE and the DisNet.
\end{enumerate}

\begin{figure*}[!t]
\centering
\includegraphics[width=\textwidth]{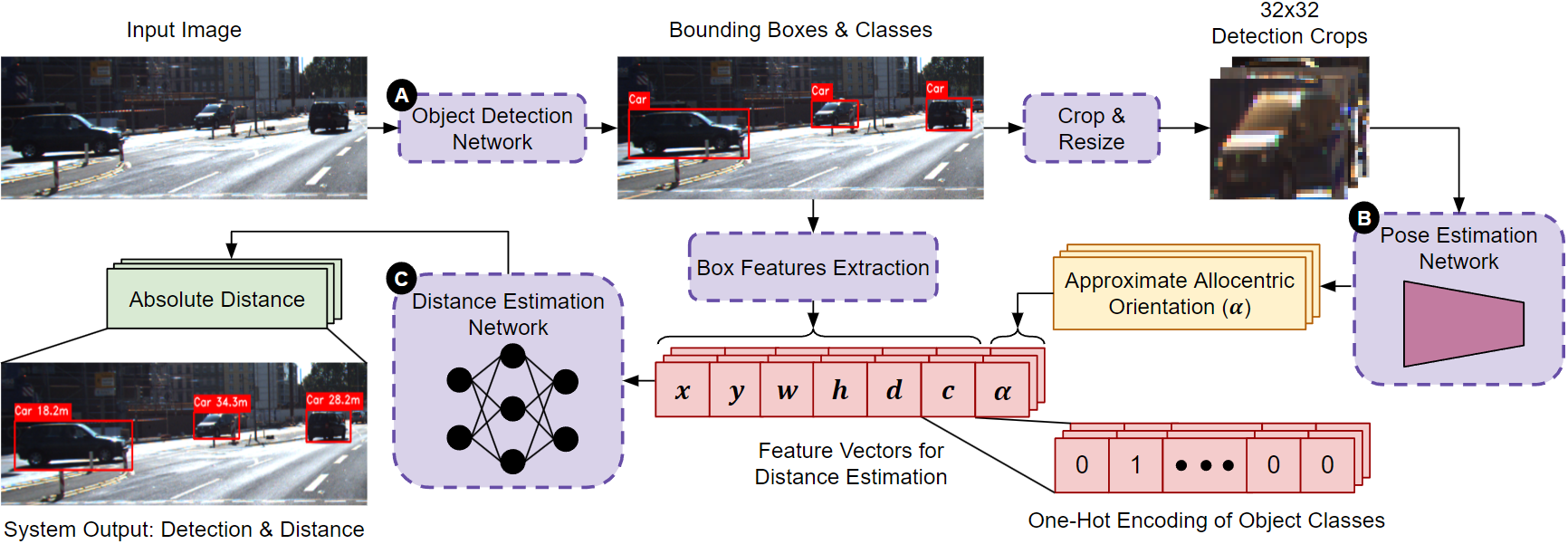}
\caption{Overview of the DECADE model, attached to an object detection network\vspace{-10pt}}
\label{mainmeth}
\end{figure*}

\section{Methodology}

We describe our detection-wise DECADE model in Figure~4. The input image is first processed by the object detector (Fig. 4\circled{A}), yielding detections, including bounding boxes and classes. These are then used to extract detection-wise features. For each detection, the image crop is processed by the Pose DNN (Fig. 4\circled{B}) to obtain the effective allocentric orientation of the detected object. This completes the feature vector for the Distance DNN (Fig. 4\circled{C}) to assign absolute distance to the detection. The full set of features and DNNs is described in the following subsections.

\subsection{Feature Analysis}
We obtain the following features from detector outputs:
\begin{itemize}
    \item Detection Class
    \item Dimensional features: width, height, diagonal
    \item Positional features: center (x,y), detection crop
\end{itemize}
The analysis of the relationship between these features and the distance of the object determines the efficacy of their inclusion in our modeling strategy. We use the KITTI 3D Object Detection \textcolor{green}{~\cite{kitti}} dataset for our feature analyses in this section and, eventually, for the evaluation of DECADE.

It is a prerequisite to include the class in the set of selected input features for any distance modeling as it allows the learning of class-wise differences: the detection instances of pedestrians vary greatly from those of cars in terms of the width to height ratio. Our analysis of the dimensional features on the approximately 40k bounding box annotations in the KITTI dataset is in agreement with the main hypothesis that closer objects appear bigger. Additionally, height is the most robust dimensional feature as it effectively addresses cases where equidistant objects may differ in their widths and diagonals, but not in height (see Figure~6a for examples). Still, we hypothesize that using the full set of available features yields the most accurate model. This is because using more features minimizes the chance that detector error in any one feature significantly impacts the accuracy of distance estimation.


\begin{figure}[h]
\centering
\includegraphics[width=\linewidth]{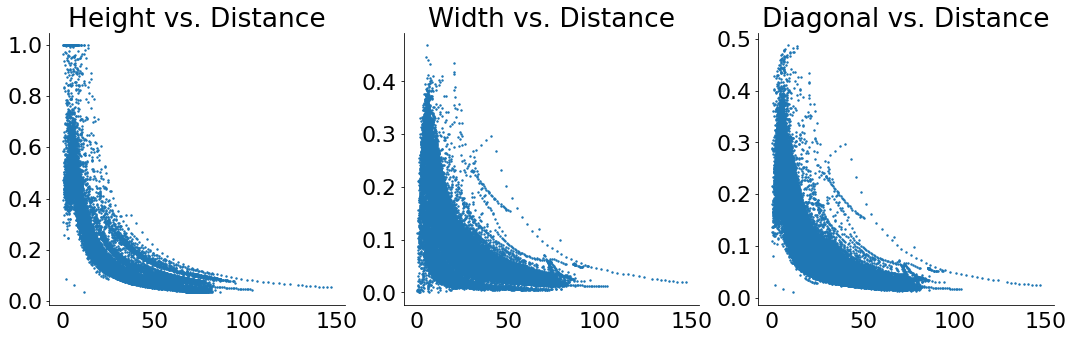}
\caption{Relationship between dimensional features (normalized by image dimensions) of bounding boxes and the object's distance from the camera~\vspace{-10pt}}
\label{weather}
\end{figure}

\begin{figure}[h]
\centering
\includegraphics[width=\linewidth]{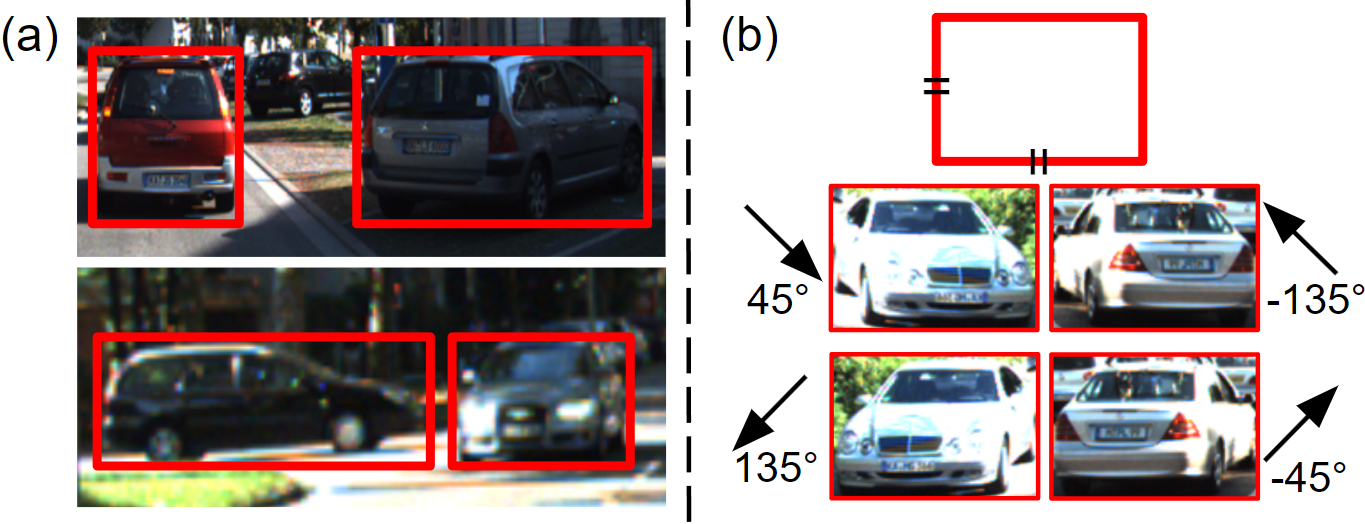}
\caption{(a) Variation in dimensional features of equidistant objects. Observe that box height remains almost identical, proving its robustness, whereas width and diagonal are affected. (b) Variation in allocentric orientation of equidistant objects with identical dimensional features, justifying the effective range reduction from [0, 360] to [0, 90] degrees~\vspace{-10pt}}
\label{weather}
\end{figure}

For accuracy optimization, we hypothesize that using the full set of available features performs better at the cost of greater system complexity. This is because using more features minimizes the chance that detector error in any one feature significantly impacts the accuracy of distance estimation.

To address the challenge of width/diagonal variation in equidistant objects, we propose the inclusion of the aforementioned positional features, in addition to the dimensional features. The difference in width/diagonal of objects in Figure~6a is due to the difference in their orientation in the road scene, not due to a difference in distance. This is why we supplement dimensional features with the allocentric orientation of the objects which can be derived from the cropped image of each detection. The effective range of this allocentric orientation can be reduced from 0-360 degrees to 0-90 degrees as we are only concerned with the effect of orientation on the visible component of the object’s width/diagonal in the 2D image (see Figure~6b).

Lastly, using information about the object’s position in the scene, i.e., center coordinates of the bounding box in combination with the allocentric orientation, it is possible to indirectly estimate the observation angle of the object with reference to the camera. Given estimates of allocentric orientation and observation angle, the object’s egocentric orientation can also be estimated \textcolor{green}{~\cite{3dboxreg}}. In this way, the use of both center coordinates and allocentric orientation adds the capability of indirectly learning egocentric orientation estimation to DECADE which further improves accuracy. This completes the feature vector presented in Figure~4.

\subsection{DNNs in DECADE}

After the object detector's (Figure~4\circled{A}) inference, DECADE leverages all the available information from the outputs to model distance for each detection using the feature set described above. We can directly extract the class, all dimensional features, and the box center coordinates from the detector output. However, estimating the allocentric orientation requires an additional Convolutional Neural Network (CNN) to process the detection crops (Figure~4\circled{B}). Thus, we process all detection crops with this pose estimation network (PoseCNN) to infer the effective allocentric orientation. This completes the feature set for the distance estimation network.

\begin{figure}[h]
\centering
\includegraphics[width=\linewidth]{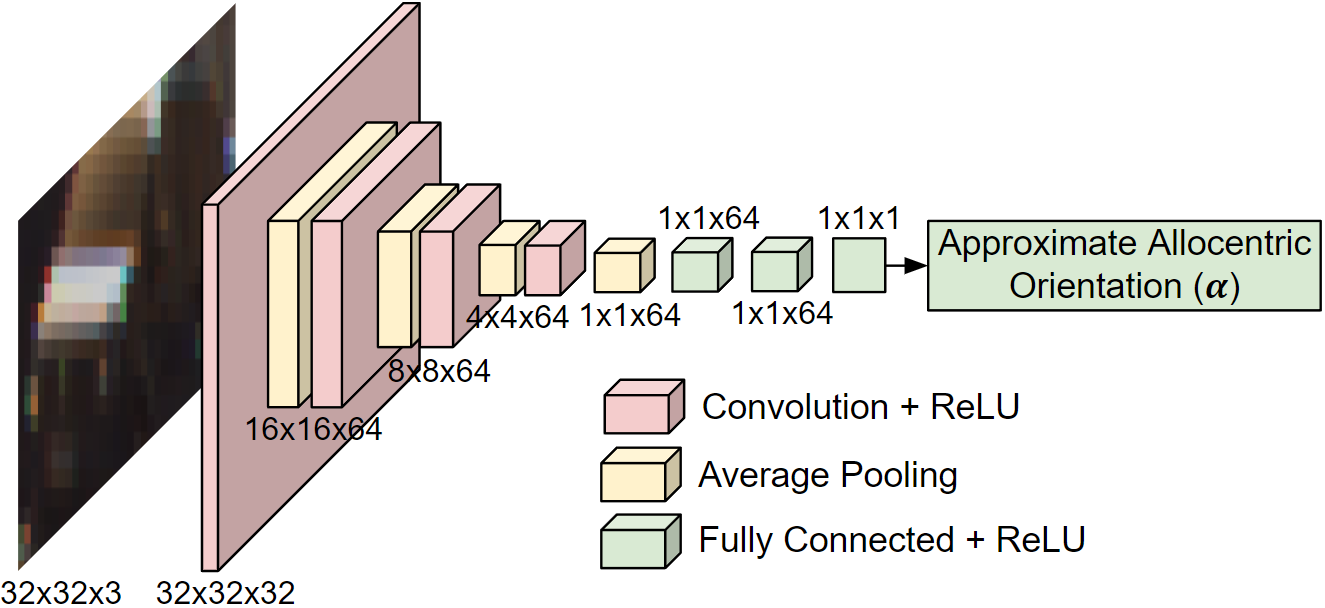}
\caption{The proposed Pose Estimation Network architecture~\vspace{-10pt}}
\label{weather}
\end{figure}

Finally, to learn a complex function that maps the feature set extracted from the detections to distance, we employ a lightweight Multi-Layer Perceptron (MLP), as shown in Figure~4\circled{C}. Our DistMLP architecture of input features feeding into 3 hidden layers of 100 neurons each, and a final output neuron, is inspired by the DisNet \textcolor{green}{~\cite{disnet}}. However, our feature set is more extensive compared to the DisNet, which only employs dimensional features and object classes.

\subsection{Adaptation to Detector Error}

We train DECADE on the bounding box annotations in the ground truth of the KITTI dataset. However, DECADE can be fine-tuned further by training on the bounding boxes yielded from the detector inference on the KITTI images directly. In this case, only supervision by the ground truth of orientation and distance is required. This ground truth can be obtained simply by matching the inference results on an image with its corresponding box annotations over a specified IoU. Since DECADE only processes detector outputs for distance estimation, this would allow it to adapt to the inaccuracy in the detector’s predictions, improving accuracy even further. Firstly, the PoseCNN adapts to the detector. Subsequently, the DistMLP can adapt by training on the effective allocentric orientation feature yielded from this PoseCNN-Adapt and the rest of the features from the detector. We expect significant accuracy differences between the end-to-end evaluation of DECADE trained on the dataset and that trained by additionally adapting to the detector.


\section{Evaluation and Discussion}

\subsection{Experimental Setup}
All experiments, except for those in Section 3.5, were conducted on the NVIDIA GeForce RTX 4090 GPU. We trained all distance estimation models with the KITTI Object Detection dataset \textcolor{green}{~\cite{kitti}}. Using the train/test split provided by the authors of Dist-YOLO \textcolor{green}{~\cite{distyolo}}, we follow their preprocessing guidelines by removing all annotations of distances less than 0 meters, and clipping the maximum to 150 meters. With 6699 images in our train, and 782 images in our test set, we use the corresponding 35450 train, and 4140 test set annotations to supervise and evaluate PoseCNN and DistMLP. Both DNNs were trained using the Adam optimizer for 250 epochs with a batch size of 64. The learning rate was set to 0.001 for PoseCNN and 0.0001 for DistMLP. We use the mean-squared loss to supervise training.  Note that the best performing DisNet was trained under the same training settings as the DistMLP, with the feature vector as specified by the author \textcolor{green}{\cite{disnet}}. The metrics used to evaluate the accuracy of distance predictions are the Mean Absolute Error (MAE), 

\[
\varepsilon_A = \frac{1}{n} \sum_{i=1}^{n} |d_i - \hat{d}_i|
\]

and the Mean Relative Error (MRE), 

\[
\varepsilon_R = \frac{1}{n} \sum_{i=1}^{n} \frac{|d_i - \hat{d}_i|}{d_i}
\]

as defined in Dist-YOLO, with $d_i$ and $\hat{d}_i$ being the groundtruth and prediction respectively \textcolor{green}{~\cite{distyolo}}. Note that the MRE is the same as the Absolute Relative Error.

\begin{figure}[h]
\centering
\includegraphics[width=\linewidth]{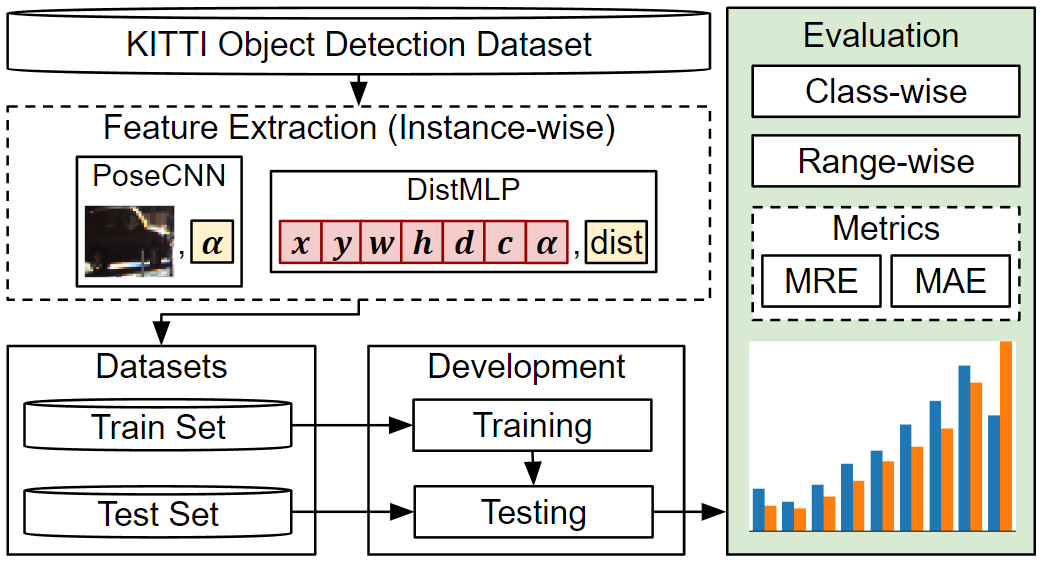} 
\caption{Evaluation pipeline for PoseCNN and DistMLP, each with its own respective datasets generated using the KITTI dataset.~\vspace{-10pt}}
\label{weather}
\end{figure}

\subsection{Evaluation on Ground Truth}
For the evaluation of PoseCNN and DistMLP on the dataset generated using the KITTI dataset, we describe our processing pipeline in Figure~8. 
Each network was trained separately, with PoseCNN learning to predict the effective allocentric orientation given the $32x32x3$ detection crop and the DistMLP learning absolute distance for a detection given its feature vector. Thus, the respective datasets extracted from KITTI include instances of this mapping. Lastly, the DNNs are evaluated against the respective test set annotations of orientation for PoseCNN and distance for DistMLP. For PoseCNN, we only use the MAE metric for evaluation. We achieve an MAE of $3.68 ^{\circ}$ for the best performing PoseCNN.

\begin{figure}[h]
\centering
\includegraphics[width=\linewidth]{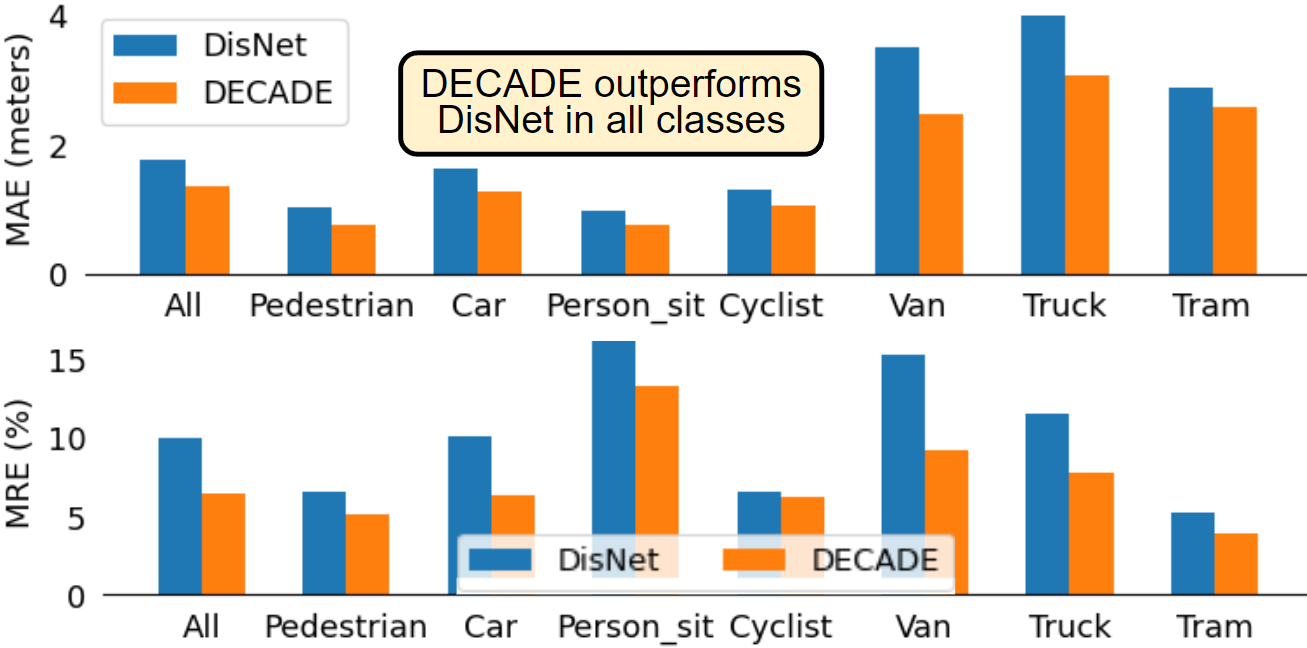} 
\caption{Class-wise evaluation of DECADE on KITTI's ground-truth test set annotations~\vspace{-10pt}}
\label{weather}
\end{figure}

\begin{figure}[h]
\centering
\includegraphics[width=\linewidth]{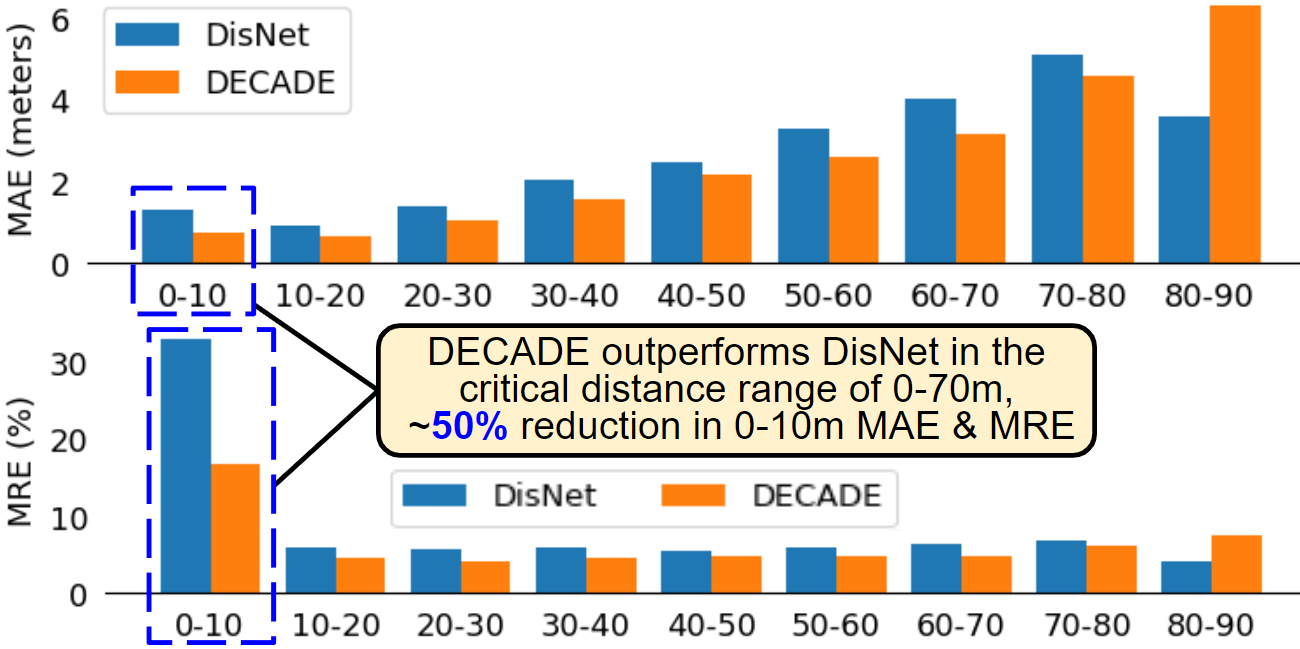}
\caption{Range-wise evaluation of DECADE on KITTI's ground-truth test-set annotations~\vspace{-5pt}}
\label{weather}
\end{figure}

In Figure~9, a class-wise accuracy comparison proves that DECADE significantly outperforms the DisNet. Over all classes, DECADE achieves $1.36$m in MAE and a $6.35\%$ MRE, compared to the DisNet's $1.76$m MAE and $9.89\%$ MRE. This is an over 22\% reduction in MAE and over 35\% in MRE. Furthermore, we not only evaluate distance estimation accuracy across the classes in the KITTI dataset, but also over the distance range of 0-90m. For an ADAS application, we believe high performance is required in the critical distance range of 0-70m. Furthermore, emphasis on model performance increases the shorter the distance is from the camera. Figure~10 shows that DECADE outperforms DisNet in this aspect as well. Most significantly, it achieves 0.75m MAE and 16.71\% MRE in the 0-10m range. Compared to the DisNet's 1.30m MAE and 33.1\% MRE, this is an over 40\% reduction in MAE and about 50\% in MRE. Note that the MRE is a highly sensitive metric at low distances. For example, a prediction of 1.1m compared against the ground truth value of 1.0m yields a 10\% relative error with only a 0.1m absolute error.

With these results, we prove that positional information is essential in the feature set for high quality distance estimation. Additionally, given an ideal object detector with no error in box location and dimensions (KITTI test-set annotations in this evaluation case), DECADE is the most accurate detection-wise distance estimation model in the current literature. Next, we evaluate our models end-to-end with the YOLOv8n to assess the distance estimation performance of the entire pipeline in a realistic scenario where the detector has the capacity for error.

\begin{figure}[h]
\centering
\includegraphics[width=\linewidth]{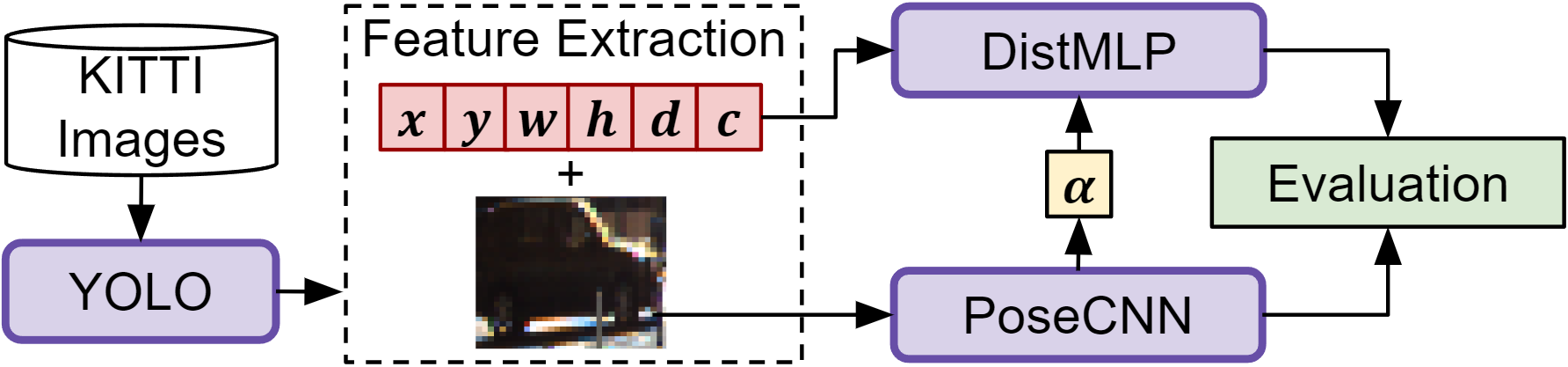} 
\caption{End-to-end evaluation pipeline for the best performing PoseCNN and DistMLP, combined with YOLO~\vspace{-10pt}}
\label{weather}
\end{figure}

\subsection{End-to-End Evaluation with YOLOv8}

For end-to-end evaluation with an object detector, we employ the YOLOv8 variants of size nano (n), small (s), and medium (m) from the open-sourced Ultralytics framework \textcolor{green}{\cite{ultralytics}}. These detectors have been trained over the KITTI dataset with the following settings: (640,200) image size, 200 epochs, batch size of 24, and 0.001 learning rate with the Adam optimizer. We present the performance evaluation results of each of these variants in Table 1.

Since YOLOv8n has lowest mAP score, it has the greatest capacity for error. Thus, we use this variant in our evaluation of the end-to-end system to examine the performance of DECADE in the worst case of highest detector inaccuracy. Figure~11 presents the overall evaluation pipeline for the end-to-end evaluation setting. We match the predictions of YOLOv8n with ground truth box annotations in the KITTI dataset over a strict IoU threshold of 0.6. This allows us to obtain target values for a comparison with DECADE's predictions for evaluation.

\begin{table}[]
\centering
\caption{Accuracy and parametric complexity of the YOLOv8 Object Detectors trained on the KITTI dataset.}
\resizebox{\columnwidth}{!}{%
\begin{tabular}{|l|l|l|ll}
\cline{1-3}
YOLOv8 variant & mAP over 0.5 IoU (\%) & Parameters (millions)  \\ \cline{1-3}
n              & 79.97               & 3.2                    \\ \cline{1-3}
s              & 83.46               & 11.2                   \\ \cline{1-3}
m              & 83.80               & 25.9                   \\ \cline{1-3}
\end{tabular}%
}
\end{table}

\begin{figure}[h]
\centering
\includegraphics[width=\linewidth]{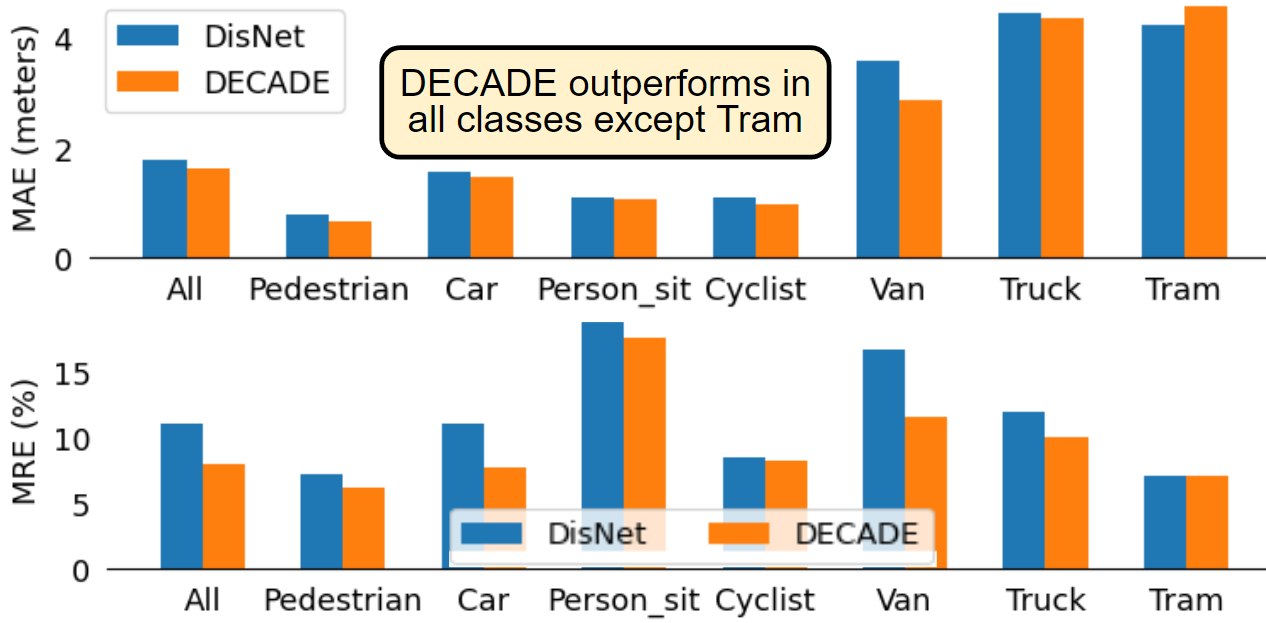}
\caption{Class-wise end-to-end evaluation of DECADE on features extracted from YOLOv8n, using PoseCNN for orientation angle}~\vspace{-10pt}
\label{weather}
\end{figure}

\begin{figure}[h]
\centering
\includegraphics[width=\linewidth]{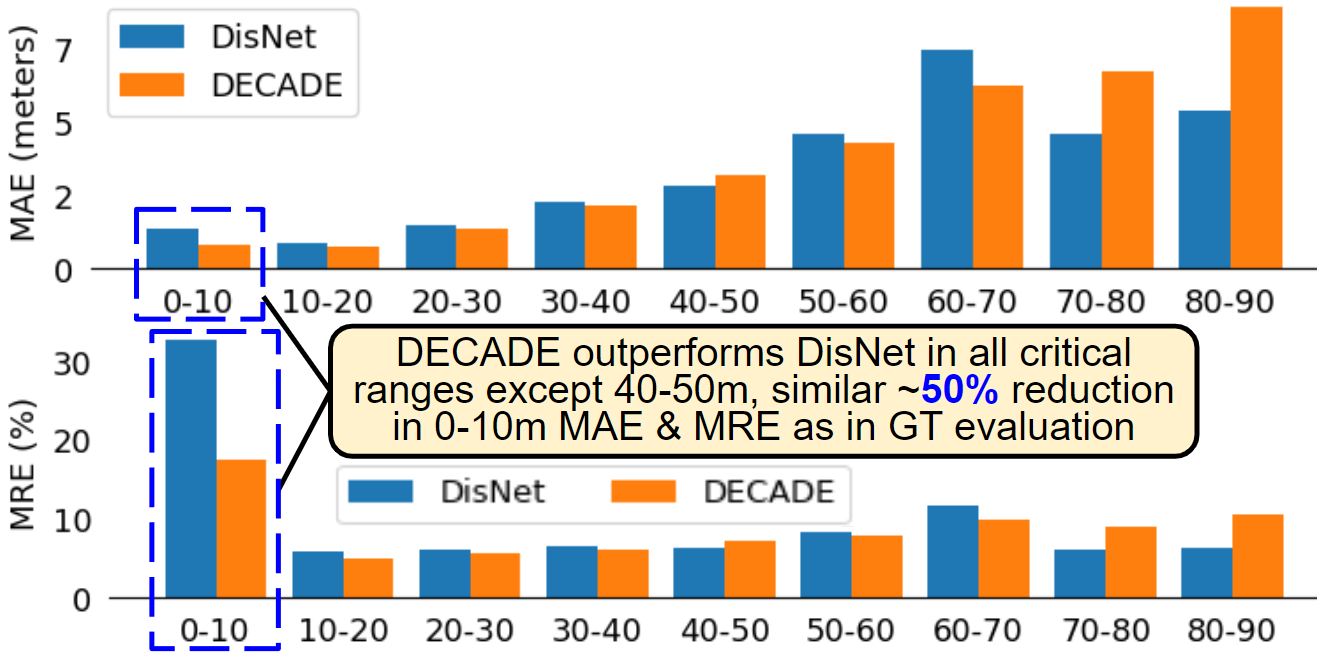}
\caption{Range-wise end-to-end evaluation of DECADE on features extracted from YOLOv8n, using PoseCNN for orientation angle}~\vspace{-20pt}
\label{weather}
\end{figure}

Firstly, PoseCNN achieves $6.45 ^{\circ}$ MAE when evaluated end-to-end with detection crops from YOLO inference. Figures~12 and 13 demonstrate that the end-to-end DECADE retains its class-wise and range-wise effective edge over the end-to-end DisNet. Here, DECADE achieves a class-wise overall MAE of 1.62m and MRE of 7.98\%, compared to the DisNet's MAE of 1.77m and 11.12\% MRE.

However, the end-to-end performance of DECADE suffers from slightly higher errors than when evaluated on ground-truth. This is especially apparent in the increase of PoseCNN MAE and in the errors for the class \textit{Tram}, where the DisNet has slightly lower MAE (4.22m vs. 4.55m) and MRE (7.04\% vs. 7.08\%). With this, we further justify the need for the adaptation scheme in Section 2.3 as it allows DECADE to adapt to detector inaccuracy, yielding more robust distances.

\begin{figure}[h]
\centering
\includegraphics[width=\linewidth]{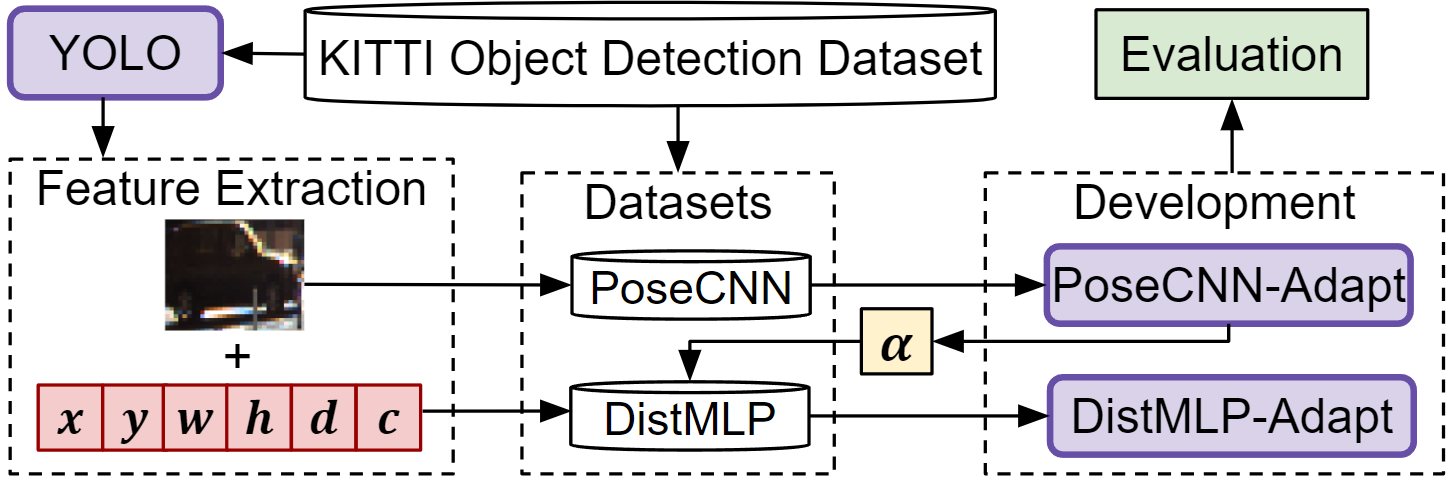} 
\caption{Evaluation pipeline for adapted networks. The datasets contain YOLO inference on KITTI, matched with annotated boxes over 0.6 IoU to get actual orientation and distance. PoseCNN adapts to YOLO, then DistMLP adapts to PoseCNN-Adapt and YOLO. The Development block consists of finetuning pretrained networks.}~\vspace{-10pt}
\label{weather}
\end{figure}


\subsection{Adapted Networks}

The evaluation pipeline for adapted networks is described in Figure~14. The best performing networks are trained further for 100 epochs, using the datasets described in the figure. The PoseCNN-Adapt achieves $3.80 ^{\circ}$, an improvement over the PoseCNN trained over KITTI only. Similarly, the DistMLP that adapts to YOLO and the PoseCNN-Adapt improves over the base network.

\begin{figure}[h]
\centering
\includegraphics[width=\linewidth]{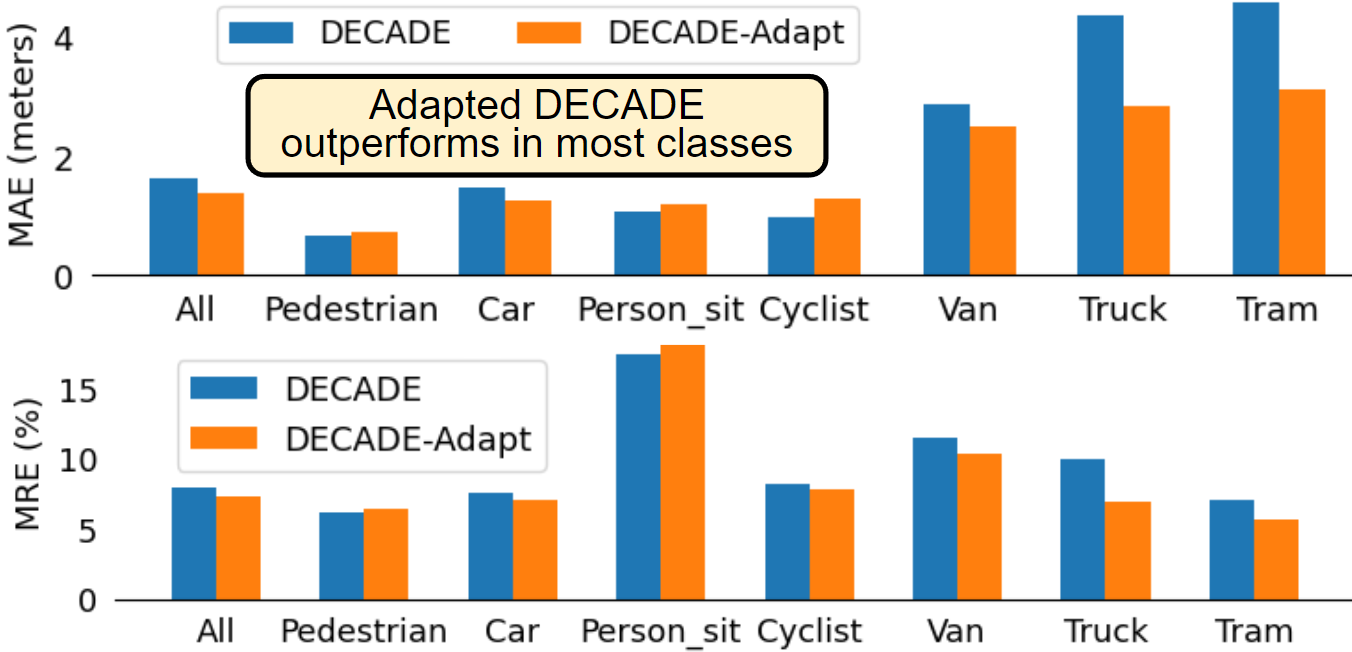}
\caption{Class-wise performance comparison of adapted networks~\vspace{-10pt}}
\label{weather}
\end{figure}

\begin{figure}[h]
\centering
\includegraphics[width=\linewidth]{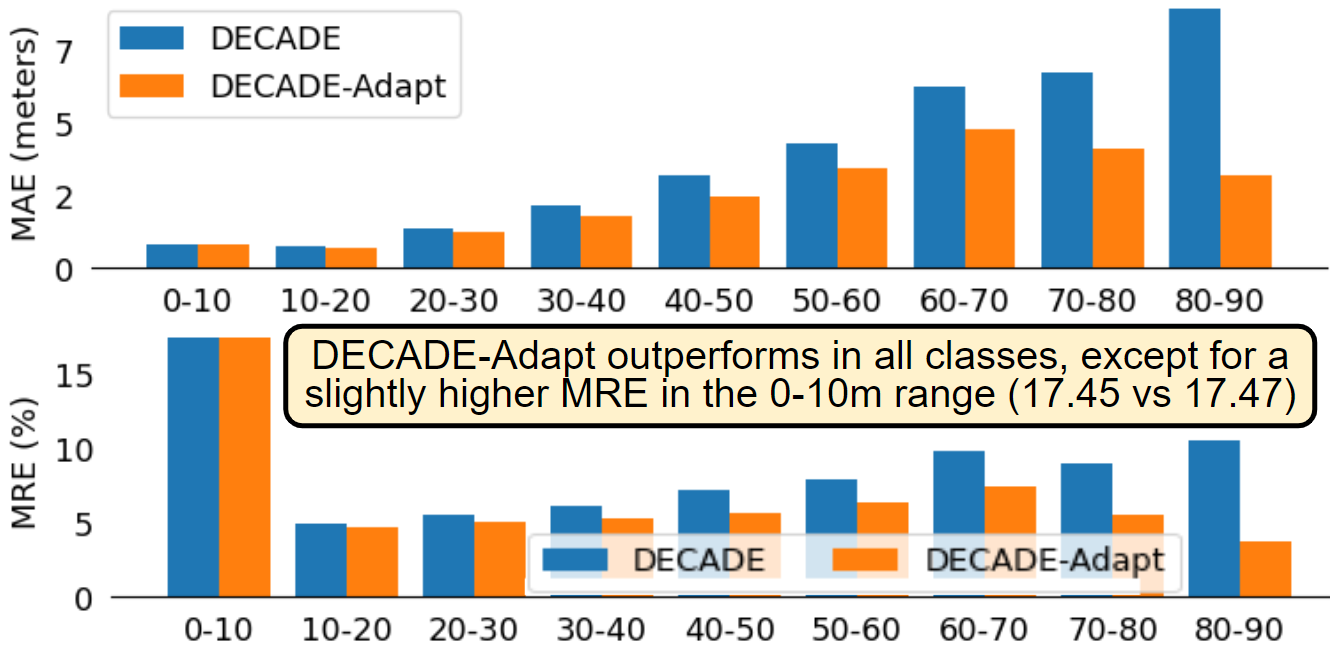}
\caption{Range-wise performance comparison of adapted networks~\vspace{-20pt}}
\label{weather}
\end{figure}

Figures 15 and 16 demonstrate that the adaptation scheme yields improvements in almost all aspects. Overall, the DECADE-Adapt achieves 1.38m MAE and 7.98\% MRE, improving over the base DECADE's end-to-end performance of 1.62m MAE and 7.38\% MRE. These results prove that adaptation yields higher performance by learning to address the errors of the specific detector employed.

\begin{table}[h]
\centering
\caption{Parametric complexity, number of FLOPs, and inference time averaged over a thousand inferences for each neural network in the object detection and distance estimation pipeline.}
\resizebox{\columnwidth}{!}{%
\begin{tabular}{|l|r|r|r|l}
\cline{1-4}
Neural Network & \multicolumn{1}{l|}{Parameters} & \multicolumn{1}{l|}{FLOPs} & \multicolumn{1}{l|}{\begin{tabular}[c]{@{}c@{}}Inference Time (ms)\end{tabular}} \\ \cline{1-4}
YOLOv8n        & 3.2M                            & 8.7B                       & 14.55                                     \\ \cline{1-4}
MonoDepth      & 14.8M                           & 11.6B                      & 25.88                                     \\ \cline{1-4}
PoseCNN        & 102.3K                          & 8.3M                       & 0.63                                      \\ \cline{1-4}
DisNet/DistMLP & 22.3K                           & 22.6K                      & 0.14                                      \\ \cline{1-4}
\end{tabular}
}
\end{table}

\begin{table}[h]
\caption{Efficiency evaluation of different systems combining specialized DNNs for object detection and distance estimation.}
\resizebox{\columnwidth}{!}{%
\begin{tabular}{|l|r|r|r|l}
\cline{1-4}
System Configuration & \multicolumn{1}{l|}{Parameters} & \multicolumn{1}{l|}{FLOPs} & \multicolumn{1}{l|}{\begin{tabular}[c]{@{}c@{}}Inference \\Time (ms)\end{tabular}}  \\ \cline{1-4}
YOLO+MonoDepth       & 18.0M                           & 20.3B                      & 40.43                                     \\ \cline{1-4}
YOLO+DisNet          & 3.2M                            & 8.7B                       & 14.69                                     \\ \cline{1-4}
YOLO+PoseCNN+DistMLP          & 3.3M                            & 8.7B                       & 15.32                                     \\ \cline{1-4}
\end{tabular}%
~\vspace{-10pt}}
\end{table}

\subsection{Evaluating the Accuracy-Efficiency Tradeoff}

Having compared the accuracy of DECADE with DisNet, it is important to discuss the parametric complexities and efficiency of both methods. Since many DNN accelerators are designed for larger networks, it is important to choose the evaluation platform wisely to not exaggerate or minimize the differences between the large YOLO and the smaller PoseCNN and DistMLP networks \textcolor{green}{\cite{scalpel, eyeriss}}. Thus, to avoid the intricacies of finding a suitable evaluation platform that offers a fair comparison between networks with significant architectural differences, we use a computer with AMD Ryzen Threadripper PRO 5975WX 32-Cores (256 GB RAM) to generate the following results, instead of a GPU. Table 2 highlights the performance characteristics of each DNN employed in our evaluation pipeline. In DECADE, the DistMLP has the same parametric complexity as the DisNet. However, the PoseCNN adds around 100k trainable parameters, 8.3M FLOPs, and 0.63ms additional inference time. 

Table 3 shows the different system configurations, i.e., combinations of neural networks to build a system for object detection and distance estimation. Observe that both the detection-wise distance estimation systems DisNet and DECADE (PoseCNN+DistMLP) are far more efficient than the pixel-wise MonoDepth, since they have less than half the parameters, FLOPs and inference time. At this point, we rule out the MonoDepth from further efficiency evaluation. 

Furthermore, in DECADE, the effect of adding PoseCNN on the overall system efficiency is almost negligible. Still, we justify DECADE's increase in complexity compared to the DisNet with an over 30\% improvement in MRE in distance estimation accuracy at the cost of only a 3\% addition to total parameters and 4\% to inference time. In the future, we will implement DNN compression techniques like pruning to further reduce the PoseCNN's complexity. Overall, Figure~17 illustrates how our research unlocks a new frontier in the accuracy-efficiency tradeoff of distance estimation models, offering significantly better accuracy at the parametric complexities of the most cost-effective methods.

\begin{figure}[h]
\centering
\includegraphics[width=\linewidth]{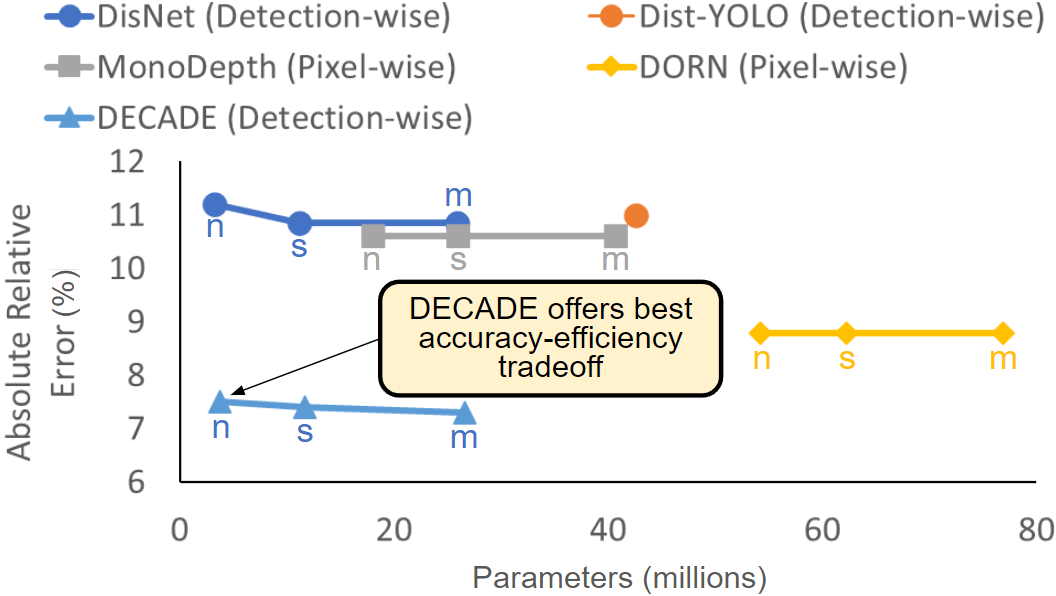}
\caption{Accuracy and parametric complexity of DECADE compared to the state-of-the-art presented in Fig. 3\vspace{-10pt}}
\label{weather}
\end{figure}

\begin{table}[h]
\caption{Latency evaluation of each component on different edge devices (smartphones), averaged over a thousand inferences. DECADE only introduces the minor additional latency of PoseCNN, when compared with DisNet.}
\resizebox{1\linewidth}{!}{%
\begin{tabular}{lrrrl}
\cline{1-4}
\multicolumn{1}{|l|}{Smartphone Model}   & \multicolumn{1}{l|}{\begin{tabular}[c]{@{}c@{}}YOLO \\(ms)\end{tabular}} & \multicolumn{1}{l|}{\begin{tabular}[c]{@{}c@{}}PoseCNN \\(ms)\end{tabular}} & \multicolumn{1}{l|}{\begin{tabular}[c]{@{}c@{}}DisNet/\\DistMLP (ms)\end{tabular}}  \\ \cline{1-4}
\multicolumn{1}{|l|}{Samsung Galaxy A70} & \multicolumn{1}{r|}{155.88}    & \multicolumn{1}{r|}{1.93}         & \multicolumn{1}{r|}{0.78}                 \\ \cline{1-4}
\multicolumn{1}{|l|}{OnePlus 8 Pro}      & \multicolumn{1}{r|}{74.87}     & \multicolumn{1}{r|}{1.12}         & \multicolumn{1}{r|}{0.47}                 \\ \cline{1-4}
\multicolumn{1}{|l|}{Samsung Galaxy S21} & \multicolumn{1}{r|}{52.67}     & \multicolumn{1}{r|}{1.06}         & \multicolumn{1}{r|}{0.44}                 \\ \cline{1-4}
\end{tabular}%
~\vspace{-10pt}}
\end{table}

Finally, Table 4 demonstrates how our evaluation extends to edge devices to assess the efficacy of the proposed solution in a mobile ADAS application. With the Samsung Galaxy S21, the overall system latency translates to around 19 FPS on the edge device. This proves that DECADE can offer state-of-the-art performance in resource constrained hardware. Note that since we have not implemented any DNN compression techniques yet, we expect significant improvements after hardware-specific optimizations.

\section{Conclusion}

In conclusion, this paper addresses the critical need for efficient yet accurate detection-wise distance estimation models in ADAS. By leveraging lightweight DNNs integrated with object detection networks, we propose a novel approach to addressing collision avoidance that significantly reduces computational complexity while maintaining high accuracy. Through our DECADE model, which extends object detectors with detection-wise distance estimation networks supplemented by pose estimation, we achieve state-of-the-art performance in accuracy (7.3\% absolute relative and 1.38m mean absolute error) while remaining cost-effective. Our contributions pave the way for the development of more accessible and effective ADAS solutions, ultimately contributing to the improvement of road safety worldwide.

\section*{Acknowledgements}
This work was partially supported by the NYUAD Center for Artificial Intelligence and Robotics (CAIR), funded by Tamkeen under the NYUAD Research Institute Award CG010, and the NYUAD Center for Interacting Urban Networks (CITIES), funded by Tamkeen under the NYUAD Research Institute Award CG001.







\bibliographystyle{IEEEtran}
\bibliography{main.bib}

\end{document}